\PassOptionsToPackage{obeyspaces}{url}
\documentclass[sigconf,screen]{acmart}

\usepackage[acm,subfig]{definition}
\usepackage{bigstrut}

\copyrightyear{2022} 
\acmYear{2022} 
\setcopyright{acmcopyright}\acmConference[KDD '22]{Proceedings of the 28th ACM SIGKDD Conference on Knowledge Discovery and Data Mining}{August 14--18, 2022}{Washington, DC, USA}
\acmBooktitle{Proceedings of the 28th ACM SIGKDD Conference on Knowledge Discovery and Data Mining (KDD '22), August 14--18, 2022, Washington, DC, USA}
\acmPrice{15.00}
\acmDOI{10.1145/3534678.3542680}
\acmISBN{978-1-4503-9385-0/22/08}

\usepackage{algorithm}
\usepackage{algpseudocode}
\annotator{yanqiao}{red}



\begin{document}

\title[Data-Efficient Brain Connectome Analysis via Multi-Task Meta-Learning]{Data-Efficient Brain Connectome Analysis \\ via Multi-Task Meta-Learning}

\author[Yi Yang, Yanqiao Zhu, Hejie Cui, Xuan Kan, Lifang He, Ying Guo, and Carl Yang]{Yi Yang$^{1,}$*, Yanqiao Zhu$^{2,}$*, Hejie Cui$^{1}$, Xuan Kan$^{1}$, Lifang He$^{3}$, Ying Guo$^{4}$, and Carl Yang$^{1,\dag}$}

\authornotetext{The first two authors made equal contribution to this work.}
\authornotetext{To whom correspondence should be addressed.}

\affiliation{%
	\institution{$^1$Department of Computer Science, $^4$Biostatistics and Bioinformatics, Emory University}
	\institution{$^2$Department of Computer Science, University of California, Los Angeles}
	\institution{$^3$Department of Computer Science and Engineering, Lehigh University}
	\country{}
}

\email{{yi.yang, hejie.cui, xuan.kan, yguo2, j.carlyang}@emory.edu, yzhu@cs.ucla.edu, lih319@lehigh.edu}

\def\authors{Yi Yang, Yanqiao Zhu, Hejie Cui, Xuan Kan, Lifang He, Ying Guo, and Carl Yang}

\begin{abstract}
Brain networks characterize complex connectivities among brain regions as graph structures, which provide a powerful means to study brain connectomes.
In recent years, graph neural networks have emerged as a prevalent paradigm of learning with structured data.
However, most brain network datasets are limited in sample sizes due to the relatively high cost of data acquisition, which hinders the deep learning models from sufficient training.
Inspired by meta-learning that learns new concepts fast with limited training examples, this paper studies data-efficient training strategies for analyzing brain connectomes in a cross-dataset setting.
Specifically, we propose to meta-train the model on datasets of large sample sizes and transfer the knowledge to small datasets.
In addition, we also explore two brain-network-oriented designs, including atlas transformation and adaptive task reweighing.
Compared to other pre-training strategies, our meta-learning-based approach achieves higher and stabler performance, which demonstrates the effectiveness of our proposed solutions.
The framework is also able to derive new insights regarding the similarities among datasets and diseases in a data-driven fashion.
\end{abstract}

\maketitle

\section{Introduction}

It has long been an enticing pursuit for neuroscience researchers and mental disorder clinicians to understand the functions and structures of human brains, which are known to be related to many complicated diseases, including bipolar disorder (BP), immunodeficiency virus infection (HIV), and Parkinson's disease (PD) \cite{Zhang:2018wo}.
In the last decade, the development of neuroimaging techniques, such as magnetic resonance imaging (MRI), functional MRI, diffusion tensor imaging (DTI), etc., provides an important source of information that facilitates the diagnosis of various brain diseases.
Based on neuroimaging data, one can build brain networks that encode brain anatomical regions as nodes and their connections as edges. This kind of data representation characterizes the complex connections among different regions of interest (ROI) and thus is of paramount research values to the study of understanding brain connectomes and brain-related diseases.

In this deep learning era, graph neural networks (GNNs) have become a popular framework for analyzing structured data \cite{Kipf:2017tc,Velickovic:2018we}. Many computational approaches based on GNNs for brain disorder analysis have been developed \cite{cui2022braingb,Ma:2017cp,Li:2021fa,zhu2022joint,cui2022interpretable,kan2022fbnetgen}.
However, it is widely known that with insufficient training data, deep neural networks are prone to overfitting and poor generalization ability, since most deep models usually have a large number of parameters to learn \cite{belkin2019reconciling,Xu:2021vr}.
On the other hand, in real clinical studies, most brain network datasets are limited in the sample size, due to the high cost of data collection, preprocessing, and annotation. For example, the BP dataset has only 52 bipolar subjects in euthymia and 45 samples of healthy controls; another widely used HIV dataset contains merely 35 HIV patients and 35 seronegative controls \cite{Zhang:2018wo}.
Therefore, under the common fully supervised learning setting, GNNs have difficulty in learning informative knowledge due to the small sizes of brain network datasets. 

Recently, to improve data efficiency, the framework of meta learning has attracted a lot of attention in many application domains, which aims at quickly adapting deep models to new tasks using a small amount of data \cite{Finn:2017wn}.
Inspired by its success, we propose to address the aforementioned data scarcity issue in brain connectome analysis by leveraging meta-learning techniques in a cross-dataset setting.
Specifically, instead of following the common supervised learning paradigm that trains a model from scratch, we firstly train the model from different source tasks using other available datasets.
Considering the multimodal nature of the brain network datasets where multiple interrelated types of connections exist among ROIs (e.g., structural and functional connections), we propose to leverage every such modality as one training task and meta-train the models using multiple training tasks.
With sufficient amount of meta-training, we have a pre-trained model that simultaneously performs well on all these training tasks, which we believe to be generic and easily transferable to new target tasks.
In the experiments, we compare the meta-learning-based approach with another popular training paradigm, namely single-task transfer learning. The results show that our proposed approach significantly improves and stabilizes the performance across a wide range of scenarios involving multiple target tasks.

Based upon the meta-learning framework, we further improve the model with brain-network-oriented designs.
At first, the datasets used for training and testing usually use different ROI atlas mappings for constructing the brain networks, resulting in different numbers and physical regions of nodes, which hinders the transferability of GNNs.
To mitigate this discrepancy, we propose to leverage a non-linear autoencoder model to preprocess all node features, such that different parcellations of brain regions in different datasets can be automatically aligned with each other.
Secondly, in the meta-training phase, different training tasks may contribute differently to the learning of generic and transferable knowledge which may limit the generalization performance. We motivate our design consideration by visualizing the relative contribution of the source tasks towards the learning on the target task, where the data-driven observation corroborates with existing clinical research. Based on our findings, we then propose an adaptive task reweighing scheme to dynamically adjust the learning rate and weight decay parameters according to the contribution of each meta-training task. Extensive experiments conducted on real-world brain network datasets verify the effectiveness of these proposed strategies. 

Overall, we summarize our contributions into three major perspectives listed below:
\begin{itemize}[wide]
    \item To the best of our knowledge, we are the first to highlight the inherent challenge of limited training samples for learning with brain network data. We formulate this problem into a data-efficient meta-learning objective where we aim for the model to quickly converge to a generic prior in a cost-effective fashion.
    \item We propose to leverage transfer learning and meta-learning strategies to pre-train a given model on available source tasks to obtain a generic and easily transferrable parameter initialization. Our methods also incorporate brain-network-oriented design considerations that effectively addresses challenges unique to brain network data.
    \item Thorough and detailed experiments have demonstrated that our proposed framework achieves a relative gain of 21.4\% and an absolute gain of 12.3\% in classification accuracy compared to the baseline of direct supervised training, and clearly outperforms other compared methods.
\end{itemize}

\section{Related Work}

\subsection{GNNs for Brain Connectome Analysis}
In recent years, graph neural networks (GNNs) have attracted broad interest due to their established power for analyzing graph-structured data \cite{Velickovic:2018we,xu2019powerful,Kipf:2017tc}. Several pioneering deep models have been devised to predict brain diseases by learning the graph structures of brain networks.
For instance, \citet{Li:2021fa} propose BrainGNN to analyze fMRI data, where ROI-aware graph convolutional layers and ROI-selection pooling layers are designed for neurological biomarker prediction.
\citet{kawahara2017brainnetcnn} design a CNN framework BrainNetCNN composed of edge-to-edge, edge-to-node, and node-to-graph convolutional filters that leverages the topological locality of structural brain networks.
The preliminary analysis of applying various existing GNN models on brain network datasets \cite{cui2022interpretable, kan2022fbnetgen, zhu2022joint} demonstrate that powerful GNNs are potentially useful for brain connectome analysis, especially when training data are relatively sufficient, which lead to significantly improved performance in tasks such as the prediction of clinical outcomes.
However, the training of powerful GNNs is extremely hard and unstable when the training data are limited, leading to poor performance and large variances, and the more complicated the GNNs are, the poorer performance and larger variances are in such settings \cite{zhu2022joint}.
In reality, the lack of training data is very common in neuroscience research and industry, especially considering specific domains and tasks. To fully unleash the power of GNNs towards the deeper modeling and understanding of brain network data, its effective training with limited supervision is of vital importance.

\subsection{Meta-Learning for Graph Classification}
Recently, meta-learning has drawn significant attention in the machine learning community since it is able to address the problem of limited training data.
There are also several attempts of meta-learning for GNN-based graph classification. For example, \citet{Chauhan2020FEW-SHOT} recognize unseen classes with limited labeled graph samples using meta-training.
\citet{https://doi.org/10.48550/arxiv.2012.06755} attempt to develop a general framework that can adapt to three-level tasks --- graph classification, node classification, and link prediction with meta-learning, but without considering the unique characteristics of brain networks.
\citet{ma2020adaptive} use the shared sub-structures between training classes and test classes to design a better meta-learning framework. However, none of the shared sub-structures can be utilized since brain networks are complete graphs.
Meta-MGNN \cite{guo2021few} proposes a self-supervised learning objective that predicts atom types for molecular datasets. However, there is no precise label for each node for prediction in brain networks.

\begin{figure*}
    \centering
    \includegraphics[width=0.88\linewidth]{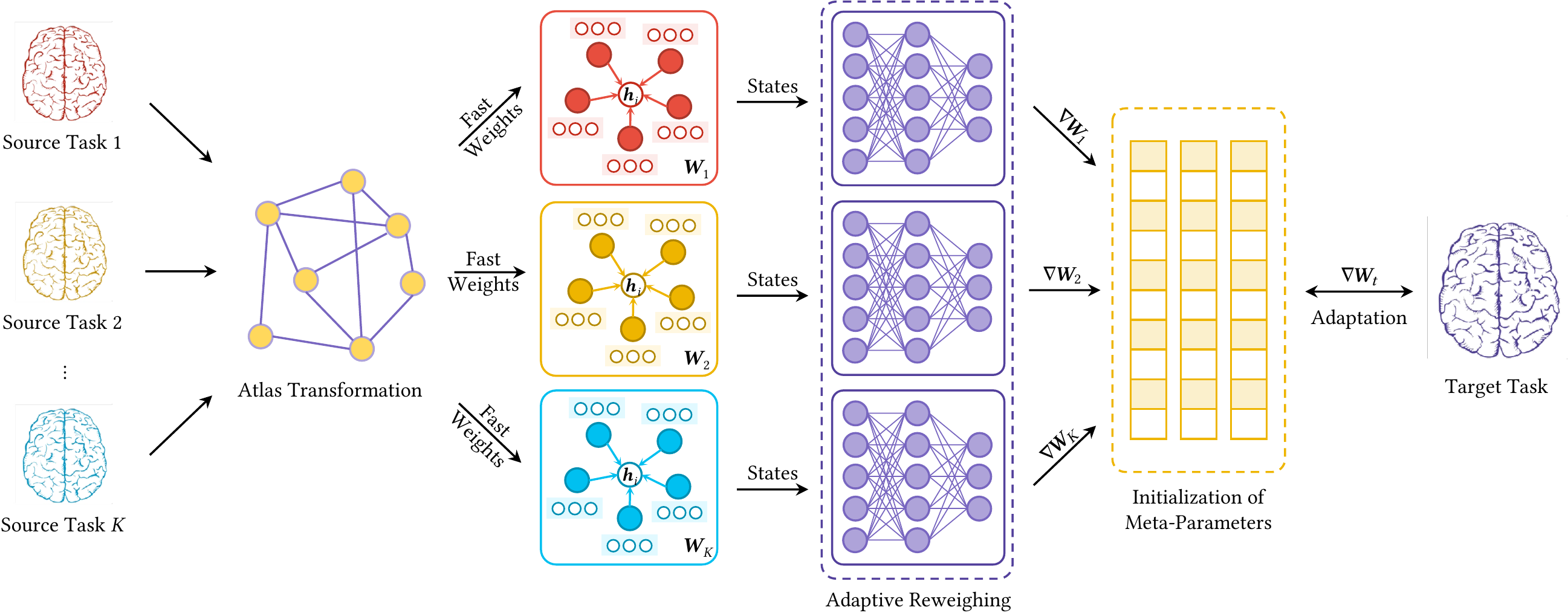}
    \caption{Overview of our proposed data-efficient learning pipeline for brain connectome analysis. First, we align data instances from various source tasks with target tasks via atlas transformation.
    Then, during the meta-training phase, we perform task-specific optimization on source tasks with a novel task adaptive reweighing scheme that dynamic adjusts the contribution of each source task.
    Finally, the learned initialization for meta-parameters is used for fine-tuning on target tasks.}
    \label{fig:overview}
\end{figure*}

\section{Preliminaries}

\subsection{Problem Definition}

We consider the problem of disease prediction with multiple brain network datasets.
For brain connectome analysis, the core is to model the interconnections between brain regions.
Formally, given a brain dataset for one specific disease $\mathcal{D} = \{(\mathcal{G}_i, y_i)\}_{i=1}^N$ containing $N$ subjects, where $\mathcal{G}_i$ represents the $i^{\text{th}}$ brain network instance and $y_i$ is its according disease label.
Each brain network object can be considered as an edge weighted graph $\mathcal{G}_i = (\mathcal{V}, \mathcal{E}_i, \bm{A}_i)$, where $\mathcal{V} = \{v_i\}_{i=1}^M$ is the node set of size $M$ describing the defined region of interests (ROIs), $\mathcal{E}_i = \mathcal{V} \times \mathcal{V}$ is the weighted edge set, and $\bm{A}_i \in \mathbb{R}^{M\times M}$ is the weighted adjacency matrix representing the connectivity among ROIs.
We define an encoder model $f(\cdot)$ parameterized by $\bm{\Theta}$ that learns a mapping relation $f_{\theta}(\mathcal{G}_i) = z_i$, where $z_i$ denotes the classification probability. 
Since each disease can be recorded in multiple datasets and each dataset can have multiple views of brain networks, we define a training task to be the prediction of one disease on a specific view of brain networks (e.g., different types of functional networks and structural networks).
In our cross-dataset multitask learning setting, we aim to train $f(\cdot)$ on a set of source tasks $\bm{S} = \{S_k\}_{k=0}^K$ to obtain $\bm{\Theta_0}$ such that the weights capture domain knowledge as sharable and generic brain structures that are useful and transferable to an unseen target task $\bm{T}$, where $\bm{S}$ and $\bm{T}$ do not necessarily concern the same type of disease.
We then aim to fine-tune $f(\cdot)$ on $\bm{T}$ such that the model can efficiently adapt to the target task optimal $\bm{\Theta}^*$ given that available training samples in $\bm{T}$ are much fewer than those in $\bm{S}$.

\subsection{Experimental Configurations}

\paragraph{Datasets.}
Our experiments use three real-world datasets of different neuroimaging modalities, which are briefly outlined below. In the experiments, the PPMI dataset is used as the meta-training dataset and we test the performance on models trained on the other two smaller datasets HIV and BP.
\begin{itemize}[wide]
    \item \textbf{Human Immunodeficiency Virus Infection (HIV).} This dataset is collected from Early HIV Infection Study in the University of Chicago. It includes both the functional magnetic resonance imaging (fMRI) and diffusion tensor imaging (DTI) for each subject. Because the original dataset was significantly imbalanced, we randomly sampled 35 early HIV patients (positive) and 35 seronegative controls for the sake of exposition. The demographic characteristics of these two sets of participants, such as age, gender, racial makeup, and educational level, are identical. We preprocess the fMRI data using the DPARSF\footnote{\url{http://rfmri.org/DPARSF/}} toolbox. We focus on 116 anatomical volumes of interest (AVOI), where a sequence of responds is extracted from them. The functional brain networks are constructed with 90 cerebral regions where each node represents a brain region and the edges are calculated as the pairwise correlation. For the DTI data, we use the FSL toolbox\footnote{\label{fsl}\url{https://fsl.fmrib.ox.ac.uk/fsl/fslwiki/}} for the preprocessing. Each subject is parcellated into 90 regions via the propagation of the automated anatomical labeling (AAL) \cite{tzourio2002automated}.
    \item \textbf{Bipolar Disorder (BP).} This dataset is collected from 52 bipolar I individuals and 45 healthy controls at the University of Chicago, including both fMRI and DTI modalities. The resting-state fMRI and DTI data were acquired on a Siemens 3T Trio scanner using a T2 echo planar imaging (EPI) gradient-echo pulse sequence with integrated parallel acquisition technique (IPAT). For the fMRI data, the brain networks are constructed using the toolbox CONN\footnote{\url{http://www.nitrc.org/projects/conn/}}, where pairwise BOLD signal correlations are calculated between the 82 labeled Freesurfer\footnote{\url{https://surfer.nmr.mgh.harvard.edu/}\label{freesurfer}}-generated cortical/subcortical gray matter regions. For DTI, same as fMRI, we constructed the DTI image into 82 regions.
    \item \textbf{Parkinson's Progression Markers Initiative (PPMI).}
    This is a restrictively public available dataset\footnote{\url{https://www.ppmi-info.org/}} to speed breakthroughs and support validation on Parkinson's Progression research. In PPMI, we consider a total of 718 subjects, where 569 subjects are Parkinson's disease patients and 149 are healthy controls. We preprocess the raw imaging using the FSL\textsuperscript{\ref{fsl}} and ANT\footnote{\url{http://stnava.github.io/ANTs/}}. 84 ROIs are parcellated from T1-weighted structural MRI using Freesurfer\textsuperscript{\ref{freesurfer}}. Based on these 84 ROIs, we construct three views of brain networks using three different whole brain tractography algorithms, namely the Probabilistic Index of Connectivity (PICo), Hough voting (Hough), and FSL. Please refer to \citet{zhan2015comparison} for the detailed brain tractography algorithms.
\end{itemize}

\paragraph{Evaluation metrics.}
In all experiments, we use two evaluation metrics: accuracy (ACC) and area under the receiver operating characteristic curve (AUC) to evaluate the classification performance of our proposed approaches. These are two extensively used evaluation metrics for disease diagnosis in medical disciplines \cite{cui2022braingb,Li:2021fa}.

\section{Data-Efficient Training Strategies}

Graph neural networks are powerful in learning representations of graph-structured objects such as brain networks.
However, under a fully supervised setting, GNN needs a relatively large-sized dataset for proper training.
With small-sized datasets like brain networks, GNNs may suffer from overfitting and fail to generalize the learned knowledge, which leads to a deterioriated performance in downstream tasks.
In this section, we study the problem of data-efficient training using multiple sources of datasets.
Specifically, given one large-sized dataset PPMI and the other two smaller-sized datasets HIV and BP, our goal is to study how to pre-train the model on PPMI (i.e. the source dataset) and use the learned knowledge to improve the performance on HIV and BP (i.e. the target datasets).

\subsection{Methodologies}
In the following, we propose to study two data-efficient training strategies for brain network analysis --- single-task supervised transfer learning and multi-task meta-learning, both of which are representative techniques in dealing with the absence of sufficient training data. 
In addition, we present two other baseline techniques, namely, direct supervised learning (without pre-training) and multi-task transfer learning (without meta-learning).

\paragraph{Method 1: Direct supervised learning (DSL)}
As a baseline investigation, we directly apply and train a randomly initialized model on a given task in a supervised fashion. The model is optimized using the binary cross-entropy objective as follows:
\begin{equation}
    \mathcal{L}_{\textsc{bce}} = -\frac{1}{|\mathcal{D}|}\sum_{(\mathcal{G}_i, y_i) \sim \mathcal{D}}y_i\log \sigma(f_\theta(\mathcal{G}_i)) + (1-y_i)\log (1-\sigma(f_\theta(\mathcal{G}_i))),
    \label{eq:pre-train}
\end{equation}
where $y_i$ stands for the ground truth label, and \(\sigma(x)= \frac{1}{1+e^{-x}}\) is the sigmoid activation function on the output logits. Specifically under this setting, the tasks are based on modalities in HIV and BP datasets, which indeed have limited training samples. We evaluate our model using the $k$-fold cross-validation framework and derive an averaged model performance.

\paragraph{Method 2: Single-task supervised transfer learning (STT)}
At first, we follow the pre-training and fine-tuning scheme in transfer learning \cite{Yang:2010dm} to distill knowledge from source task to target task in a sample-efficient way.
This framework consists of two consecutive phases: pre-training and fine-tuning.
Specifically, we first train the encoder model in a supervised manner on the source task and apply the model weights to train another encoder on the target task.

In the first pre-training phase, we train the model on the PPMI dataset using the objective described in Eq. (\ref{eq:pre-train}).
Then, in the fine-tuning phase, the trained weights \(\bm{\Theta}_0\) are used to initialize another encoder model. This model is then fine-tuned on the target task from HIV and BP datasets with the same objective function as Eq. (\ref{eq:pre-train}).
Since the model has already learned generic knowledge underneath the source task, we use a smaller learning rate to optimize the model in the fine-tuning phase. We summarize this method in Algorithm~\ref{algo:transfer}.
Note that although PPMI contains three structural views, here we define only one view as the source task since in the pre-training phase, the model is trained based on one unified objective function and cannot distinguish between multiple tasks, if they are arbitrarily grouped together.

\paragraph{Method 3: Multi-task supervised transfer learning (MTT)}
Pre-training on a singular source task is vulnerable to the inherent risk of information loss during transfer learning since the knowledge gaps among source and target domains are not readily quantifiable.
This motivates us to train a model that is initialized on some shared knowledge in multiple source tasks when they are available, such that the fine-tuning performance is not conditioned upon any particular knowledge inconsistencies from a source and target pair.

As an immediate solution, we extend STT into a multi-task setting by expanding the pre-training phase into simultaneously co-learning over multiple source objectives. That is, we regard each modality from the dataset as an individual task and our model now learns over multiple modalities.
To this end, we formulate all tasks into a distribution, where during pre-training, the model is trained on several objectives sampled from the task distribution, hence the name ``multi-task'' for this method.

Specifically for each pre-training iteration, we optimize our model parameters on a merged objective function given by
\begin{equation}
    \mathcal{L}_{\text{MTT}} = \sum_{\tau \sim \bm{S}}\mathcal{L}_{\textsc{bce}}(\tau),
    \label{eq:combine}
\end{equation}
which takes the sum over the binary cross-entropy objectives as given in Eq. (\ref{eq:pre-train}) on all source tasks. For an efficient computation, each iteration processes a mini-batch of data sampled from the source dataset. The learned weights are then used in the fine-tuning phase on the target task following the conventional supervised learning procedure.

\paragraph{Method 4: Multi-task meta-learning (MML)}
Meta-learning aims at learning a meta model that is capable of generalizing over a variety of source objectives and can quickly adapt to an arbitrary unseen task. Different from MTT, meta-learning aims at finding an optimal model initialization that enables similarly good performance on multiple pre-training tasks rather than directly combining individual models that are good for each pre-training task through averaging the model weights. This means that meta-learning can achieve better generalization, allowing efficient adaptation to unseen objectives through minimizing the risk of over-fitting the model to outperform on certain tasks while under-perform on others, which is a typical underlying concern of MTT.

Based on such intuition, we follow the widely adopted model-agnostic meta-learning (MAML) \cite{Finn:2017wn} method in our brain network learning framework. According to \citet{raghu2019rapid}, MAML is characterized by two iconic features: (1) rapid learning and (2) feature reuse, which also refers to the outer-loop update and inner-loop adaptation.
Specifically, the model is first separately trained on each objective using fast weights during the inner loop function, then the meta parameters of the model are updated by evaluating the loss against the adapted fast weights via the outer-loop module.
In other words, the model is optimized by updating on the second-order Hessian of the parameters, which leads to quicker convergence since the optimizer incorporates the additional curvature information of the loss function that helps estimate the optimal step-size along the optimization trajectory \cite{tan2019review}.
This effectively reduces the number of training iterations required to achieve a generic model. In addition, the feature reuse inner-loop performs task-specific adaptation, which results in the meta-initialization to be an informative approximation to every task.
Due to the fact that the meta-trained model does not pertain to any particular task knowledge, such initialization is therefore non-over-fitting and generically applicable to any unseen target tasks.

To be specific about our pipeline design, in the first meta-training phase, we randomly draw \(n\) training tasks with a support set (used in inner-loop) and a query set (used in outer-loop) each containing \(k\) samples from the pool of training datasets. 
Then, given the encoder model, we update the fast weights of the parameters using the common classification objective for every training task.
After training the model on all tasks, we update the meta parameters, i.e. model initialization in our case.
Thereafter, in the meta-test phase, we perform the conventional classification procedure on the target data. We summarize this method in Algorithm \ref{algo:meta}.

\begin{algorithm}[t]
    \centering
    \caption{Single-task supervised transfer learning (STT)}\label{algo:transfer}
    \begin{algorithmic}[1]
        \State \textbf{Input:} \text{pre-train task $S$, fine-tune task $T$, encoder $f(\theta)$}
        \State \textbf{Require:} \text{$\alpha$: learning rate hyperparameter}
        \State \text{Randomly initialize $\theta$}
        \State $\triangleright$ Pre-training phase
        \While{not done}
        \State \text{Evaluate the gradient} $\nabla_\theta \mathcal{L}_Sf(\theta)$
        \State Update parameters with SGD: $\theta \gets \theta - \alpha \nabla_\theta \mathcal{L}_Sf(\theta)$
        \EndWhile
        \State $\triangleright$ Fine-tuning phase
        \State Split $T$ into $T_{\text{train}}$ and $T_{\text{eval}}$ into $K$ folds
        \For{split \textbf{in} $K$ folds}
        \State Get split-specific parameters $\hat{\theta} \gets \theta$
        \While{not done}
        \State \text{Evaluate the gradient} $\nabla_{\hat{\theta}} \mathcal{L}_{T_{\text{train}}}f(\hat{\theta})$
        \State Update parameters with SGD $\hat{\theta} \gets \hat{\theta} - \alpha \nabla_{\hat{\theta}} \mathcal{L}_{T_{\text{train}}}f(\hat{\theta})$
        \EndWhile
        \State \text{Evaluate} ACC, AUC from $f_{\hat{\theta}}(T_{\text{eval}})$
        \EndFor
    \end{algorithmic}
\end{algorithm}

\begin{algorithm}[t]
    \centering
    \caption{Multi-task meta-learning (MTT)}\label{algo:meta}
    \begin{algorithmic}[1]
        \State \textbf{Input:} \text{meta-train task pool $S_\tau$, meta-test task $T$, encoder $f(\theta)$}
        \State \textbf{Require:} \text{$\alpha$, $\beta$: learning rate hyperparameters}
        \State \text{Randomly initialize $\theta$}
        \State $\triangleright$ Meta-training phase
        \While{not done}
        \For{each task $\tau_i$ \textbf{in} $S_\tau$}
        \State Sample $k$ datapoints $\mathcal{D}_i$ from $\tau_i$
        \State \text{Evaluate the gradient} $\nabla_\theta \mathcal{L}_{\mathcal{D}_i}f(\theta)$
        \State Compute the adapted parameters $\theta'_i \gets \theta - \beta \nabla_\theta \mathcal{L}_{\mathcal{D}_i}f(\theta)$
        \State Sample another set of datapoints $\mathcal{D}'_i$ from $\tau_i$
        \EndFor
        \State Update parameters $\theta \gets \theta - \alpha \nabla_\theta \sum_{\mathcal{D}'_i, \theta'_i \sim S_\tau} \mathcal{L}_{\mathcal{D}'_i}f(\theta'_i)$
        \EndWhile
        \State $\triangleright$ Meta-test phase
        \State Perform $k$-fold evaluation on target tasks
    \end{algorithmic}
\end{algorithm}

\begin{table*}[t]
  \centering
  \caption{Performance comparison of our proposed methodologies and baselines in terms of area under the ROC curve (AUC) and accuracy (ACC). The best performing model is highlighted in boldface.}
	\begin{tabular}{ccccccccccc}
	\toprule
	\multirow{2.5}{*}{Encoder} & \multirow{2.5}{*}{Dataset} & \multirow{2.5}{*}{Modality} & \multicolumn{2}{c}{{DSL}} & \multicolumn{2}{c}{{STT}} & \multicolumn{2}{c}{{MTT}} & \multicolumn{2}{c}{{MML}} \\
\cmidrule(lr){4-5} \cmidrule(lr){6-7}  \cmidrule(lr){8-9} \cmidrule(lr){10-11}    &       &       & {AUC} & {ACC} & {AUC} & {ACC} & {AUC} & {ACC} & {AUC} & {ACC} \\
	\midrule
	\multirow{4}[2]{*}{{BrainNetCNN}} & \multirow{2}[1]{*}{BP} & fMRI  & 0.50{\small ±0.13} & 0.51{\small ±0.15} & 0.55{\small ±0.07} & 0.56{\small ±0.08} & \multicolumn{1}{l}{0.56{\small ±0.09}} & \multicolumn{1}{l}{0.56{\small ±0.11}} & 0.57{\small ±0.10} & 0.57{\small ±0.07} \\
		  &       & DTI   & 0.47{\small ±0.16} & 0.49{\small ±0.14} & 0.53{\small ±0.11} & 0.54{\small ±0.12} & \multicolumn{1}{l}{0.54{\small ±0.07}} & \multicolumn{1}{l}{0.54{\small ±0.09}} & 0.55{\small ±0.13} & 0.56{\small ±0.08} \\
		  \cmidrule(lr){2-11}
		  & \multirow{2}[1]{*}{HIV} & fMRI  & 0.60{\small ±0.15} & 0.59{\small ±0.13} & 0.66{\small ±0.14} & 0.65{\small ±0.10} & \multicolumn{1}{l}{0.66{\small ±0.13}} & \multicolumn{1}{l}{0.66{\small ±0.11}} & 0.67{\small ±0.12} & 0.67{\small ±0.09} \\
		  &       & DTI   & 0.54{\small ±0.16} & 0.53{\small ±0.15} & 0.60{\small ±0.09} & 0.60{\small ±0.09} & \multicolumn{1}{l}{0.60{\small ±0.10}} & \multicolumn{1}{l}{0.60{\small ±0.12}} & 0.57{\small ±0.11} & 0.61{\small ±0.14} \\
	\midrule
	\multirow{4}[2]{*}{{GAT}} & \multirow{2}[1]{*}{BP} & fMRI  & 0.51{\small ±0.13} & 0.52{\small ±0.16} & 0.57{\small ±0.07} & 0.58{\small ±0.05} & 0.59{\small ±0.10} & 0.59{\small ±0.07} & 0.61{\small ±0.07} & 0.60{\small ±0.09} \\
		  &       & DTI   & 0.50{\small ±0.09} & 0.50{\small ±0.13} & 0.53{\small ±0.08} & 0.54{\small ±0.10} & 0.51{\small ±0.06} & 0.55{\small ±0.08} & 0.55{\small ±0.08} & 0.57{\small ±0.05} \\
		  \cmidrule(lr){2-11}
		  & \multirow{2}[1]{*}{HIV} & fMRI  & 0.61{\small ±0.15} & 0.61{\small ±0.14} & 0.65{\small ±0.07} & 0.66{\small ±0.11} & 0.66{\small ±0.09} & 0.68{\small ±0.06} & 0.68{\small ±0.10} & 0.69{\small ±0.08} \\
		  &       & DTI   & 0.56{\small ±0.17} & 0.55{\small ±0.15} & 0.61{\small ±0.07} & 0.60{\small ±0.08} & 0.62{\small ±0.09} & 0.61{\small ±0.10} & 0.64{\small ±0.09} & 0.62{\small ±0.12} \\
	\midrule
	\multirow{4}[2]{*}{{GCN}} & \multirow{2}[1]{*}{BP} & fMRI  & 0.55{\small ±0.11} & 0.54{\small ±0.14} & 0.59{\small ±0.12} & 0.58{\small ±0.13} & 0.61{\small ±0.10} & 0.60{\small ±0.11} & \textbf{0.62{\small ±0.08}} & \textbf{0.62{\small ±0.10}} \\
		  &       & DTI   & 0.51{\small ±0.12} & 0.52{\small ±0.11} & 0.52{\small ±0.10} & 0.54{\small ±0.12} & 0.55{\small ±0.09} & 0.56{\small ±0.14} & \textbf{0.59{\small ±0.07}} & \textbf{0.58{\small ±0.11}} \\
		  \cmidrule(lr){2-11}
		  & \multirow{2}[1]{*}{HIV} & fMRI  & 0.63{\small ±0.18} & 0.64{\small ±0.12} & 0.65{\small ±0.14} & 0.68{\small ±0.15} & 0.67{\small ±0.12} & 0.68{\small ±0.11} & \textbf{0.69{\small ±0.10}} & \textbf{0.70{\small ±0.09}} \\
		  &       & DTI   & 0.60{\small ±0.12} & 0.58{\small ±0.13} & 0.61{\small ±0.11} & 0.60{\small ±0.12} & 0.63{\small ±0.13} & 0.63{\small ±0.15} & \textbf{0.65{\small ±0.12}} & \textbf{0.64{\small ±0.13}} \\
	\bottomrule
	\end{tabular}
\label{tab:fullcomp}
\end{table*}

\subsection{Empirical Evaluation}
\label{subsec:trainingeval}
We conduct comparative studies to evaluate the above two strategies.
For STT, we take the three modalities from the PPMI dataset as separate pre-training tasks.
For MTT and MML, we consider each modality from the PPMI dataset as individual source task, and we merge the three view-independent tasks to build a cohort source task pool.

\subsubsection{Implementation details.}
In our experiments, we employ three representative brain network encoders in open literature: (1) BrainNetCNN \cite{KAWAHARA20171038}, (2) GAT \cite{Velickovic:2018we}, and (3) GCN \cite{Kipf:2017tc}.
The GAT encoder is composed of 4 graph attention layers with hidden dimensions of 32, 32, 32, 8.
Similarly, the GCN encoder has 4 graph convolutional layers of hidden dimensions of 32, 32, 32, 8. For BrainNetCNN, each convolutional layer is followed by the Leaky ReLU \cite{Maas13rectifiernonlinearities} activation. For GAT and GCN, the ReLU \cite{DBLP:conf/icml/NairH10} activation is deployed. A final multilayer perceptron (MLP) is attached to all models as the linear classifier. 

Regarding feature selection, we follow the practice proposed in \citet{cui2021positional} and regard the row-wise vector from the weighted adjacency matrix as input node feature. That is, the GNNs are smoothing over the original connectivity signals from input graph instances during the learning process.

For STT, MTT, and MML, we pre-train or meta-train the encoders for 150 epochs with learning rates set to 0.001 and weight decay coefficient to 0.0001.
We update the model parameters using the popular Adam \cite{DBLP:journals/corr/KingmaB14} optimizer with a cosine annealing learning rate scheduling \cite{DBLP:conf/iclr/LoshchilovH17} that sets the minimum learning rate to 0.0001.
For downstream fine-tuning and the DSL setting, the model is evaluated on the target dataset (i.e. BP and HIV) using 5 fold cross validation with each split containing approximately 80\% for training and 20\% for testing. The model is fine-tuned for 200 epochs with the same hyperparameter setup.
We report the model performance in averaged fashion along with standard deviations.

\subsubsection{Results and analysis.}
We summarize the performance on target datasets with our proposed methods in Table \ref{tab:fullcomp}. Specifically for STT, we present our performance with Hough voting in PPMI as the pre-train source task.
All data reflected in the table has passed a significant test with $p \leq 0.05$ and the best overall performance is highlighted in bold.

In general, it can be observed that STT reports consistent improvement over the baseline DSL where it brings an average of 4\% absolute gain in the AUC scores and 5\% absolute gain in the ACC scores. The reduced performance variance in STT also suggests that the pre-trained initialization improves model robustness with respect to downstream targets.
MTT, on the other hand, reports a 2\% absolute gain in ACC and 1\% gain in AUC over STT which demonstrates the effectiveness of capturing shared knowledge on the entire source dataset for helping target adaptation. 
Finally, MML brings an average of 3\% absolute gain in ACC score and 2\% gain in AUC score over MTT.
In addition, the target performance in MML is shown to have smaller variance, suggesting that the meta model is less sensitive towards different data splits.
This is because meta learning improves the generalizability of model initialization, enabling quicker adaptation on both source and target tasks. 
In summary, through extensive empirical studies, we show that the meta-learning strategy brings consistent and significant improvement for brain connectome analysis.
In addition, GCN performs the best among the three proposed encoders across the experiments which suggests its superiority in processing noisy signals and efficiency in feature extraction of brain networks.
Due to the outstanding performance of GCN, it will be our primary model used to perform subsequent experiments. 

\section{Incorporating Brain Network Oriented Techniques}

Unlike conventional graph-structured datasets, brain networks have some unique properties.
In this section, we first identify two challenges concerning learning with brain networked data.
Accordingly, we present two design considerations to address these two challenges.

\subsection{Consideration 1: Atlas Transformation}

\subsubsection{Challenges.}
For brain network data, ROI templates describe the mapping relationship between nodes and brain atlas. Once the template is chosen, all graphs in a dataset share the same amount of nodes and their physical meanings.
In our cross-dataset setting, considering that the source and target datasets are based on different templates, it is difficult to directly transfer the learned knowledge from source to target datasets due to the misalignment of nodes and dimensions in the graphs.
Although GNNs are capable of handling input graphs of varied sizes, the model is essentially learning predictive signals regarding the structures of local subgraphs \cite{Li:2018wc}, and thus simply transferring the model parameters without manipulating data-level correspondence may lead to a significant loss of information.
Note that we may directly convert different atlas to a unified one through manual mapping. However, finding all such mappings exhaustively is costly and demands tremendous expert efforts because the mapping varies across different pairs of atlas and there is often a lack of ground truth.

\subsubsection{Solutions.}
In the following, we resort to automatic learning of the atlas transformation in a data-driven fashion and present a simple solution to deal with dimensional incompatibility across different datasets.
Specifically, we propose to study the following three strategies that transform different atlas templates into one unified space:
\begin{itemize}[wide]
    \item \textbf{Zero padding.} This is a naive approach. Given source and target data with different graph dimensions, we extend the smaller scaled graphs to the same dimension as their larger counterparts by padding the graph adjacency with zero entries. Although this method is able to preserve the original graph construction and connected components, we are assuming that, semantically, a smaller scaled graph implies missing ROIs under a shared template. However, due to the absence of exact ROI template conversion among datasets, such assumption is highly vulnerable towards the risk of  forcefully altering graph semantics.
    \item \textbf{Learnable linear projection (LP).} To preserve information during the conversion, we then propose to transfer the atlas with a learnable function. Specifically, we attach a projection head at the beginning of the encoding pass that transforms the original dimension $N$ into a target dimension $M$. We characterize this projection head by implementing a learnable matrix $\bm{W} \in \mathbb{R}^{N \times M}$, which is jointly trained with the GNN encoder. Although a linearly projected output is able to preserve graph semantics to a guaranteed extent, the method is prone to unstable convergence because the output of the projection head changes at every training iteration for the same data instance, which prevents the encoder to learn from a fixed representation.
    \item \textbf{Autoencoding (AE).} To address the underlying deficiency in jointly optimizing the learnable projection matrix due to unstable training, we introduce an autoencoding technique that transforms source data into a target dimension with fixed representation in an unsupervised fashion. In particular, we construct a parameterized mapping function $f(\lambda)$ that encodes an input $\bm{X} \in \mathbb{R}^{N \times N}$ into a bottleneck output $\bm{X}' \in \mathbb{R}^{M \times M}$. To optimize the function parameter $\lambda$, we first decode the bottleneck output by passing through the inverse function $f^{-1}_{\lambda}(\bm{X}') = \hat{\bm{X}} \in \mathbb{R}^{N \times N}$, and calculate a reconstruction error with respect to the original input. Specifically, the reconstruction error is given by the MSE loss expressed as 
    \begin{equation}
    	\mathcal{L}_\text{MSE}(\bm{X}, \hat{\bm{X}}) = \frac{1}{n}\sum_{(i,j)\in [n]^2}(x_{ij} - \hat{x}_{ij})^2.
        \label{eq:ae}
    \end{equation}
    Using gradient-descent-based optimization techniques, the function parameter vector $\lambda$ at each time step $t$ is updated by the rule formulated as 
    \begin{equation}
    	\lambda^t = \alpha \lambda^{t-1} - \beta \nabla \mathcal{L}_{\lambda^{t-1}}(\bm{X},\hat{\bm{X}}),
    \end{equation}
    where $\alpha$ and $\beta$ denotes the weight decay and learning rate hyperparameter. Thus, we can separately train the projection function and the GNN encoder. Specifically, after training AE with Eq.~(\ref{eq:ae}), we extract the bottleneck output from the optimized projection function as our transformed data. The autoencoding procedure provides a simple yet effective solution towards information preservation as well as enabling the encoder to learn from a fixed input signals. 
\end{itemize}

It should be noted that, GNN models are naturally permutation invariant when performing graph classification tasks, as long as the graph-level embedding pooling operator is permutation invariant, such as the \textit{sum} operator we use in this work \cite{xu2019powerful, zhu2021transfer}. The misalignment due to different atlas is thus only a problem for node features, since we use the connection profiles as node features as suggested in \citet{cui2022braingb}. Our autoencoding procedure finds a latent space with limited dimensions that best describe the original connection profiles, which we believe are more transferable across tasks and datasets, and we observe clear performance gains in the experiments. However, we are aware that training the autoencoders separately for different tasks may still lead to misaligned latent spaces, which can be potentially handled by ordering the dimensions based on feature variances following the spirit of PCA \cite{abdi2010principal}. This is left as a future direction.

\subsubsection{Empirical evaluation.}
For our experimental setup, we align our meta-training data dimension with that of our target task. That is we choose $M$ for each meta learning objective based on the given meta test datasets.
This is because we do not need to perform transformation again on the target datasets.
Also, as the graph embeddings resulting from linear projection have preserved semantics of the source dataset, it does not alter the task distribution involved in the meta-learning framework, which allows us to directly compare the experimental results with earlier experiments.

We summarize our empirical comparison on the listed atlas transformation techniques in Table \ref{tab:atlas-mapping}.
The default method of zero padding fails to consider graph structures and ROI semantics, and thus results in the worst performance. The learnable linear projection of feature matrix presents an improvement over zero padding but suffers from high variance in performance data. This instability can be explained by the dynamic projection of input features which discourages the encoder to learn fixed input information. The non-linear autoencoding presents the best overall performance in which we observe a 2\% improvement margin over zero padding in the BP dataset and a 4\% improvement in the HIV dataset. In addition, the variance of the data distribution of the performance using autoencoding is also reduced by an average of 4\% margin compared to learnable linear projection technique. Hence, we have shown, through empirical examinations, that simple autoencoder architecture presents insightful solutions towards the challenges of data-level discrepancies of brain atlas. 

\begin{table}
  \centering
  \caption{Performance with three different atlas transformation techniques.}
  \label{tab:atlas-mapping}
  \resizebox{\linewidth}{!}{
    \begin{tabular}{cccccccc}
    \toprule
    \multirow{2.5}{*}{Dataset} & \multirow{2.5}{*}{Modality} & \multicolumn{2}{c}{Zero Pad} & \multicolumn{2}{c}{LP} & \multicolumn{2}{c}{AE} \\
    \cmidrule(lr){3-4} \cmidrule(lr){5-6} \cmidrule(lr){7-8}
    & & AUC & ACC & AUC & ACC & AUC & ACC \\
    \midrule
     \multirow{2}{*}{BP}    & fMRI & 0.62\tiny{±0.08} & 0.62\tiny{±0.10} & 0.62\tiny{±0.13} & 0.63±\tiny{0.12} & \textbf{0.63\tiny{±0.09}} & \textbf{0.64\tiny{±0.09}} \\
     & DTI & 0.59\tiny{±0.07} & 0.58\tiny{±0.11} & 0.59\tiny{±0.09} & 0.60\tiny{±0.14} & \textbf{0.60\tiny{±0.04}} & \textbf{0.61\tiny{±0.10}} \\\midrule
     \multirow{2}{*}{HIV}    & fMRI & 0.69\tiny{±0.10} & 0.70\tiny{±0.09} & 0.71\tiny{±0.13} & 0.70\tiny{±0.11} & \textbf{0.73\tiny{±0.10}} & \textbf{0.72\tiny{±0.08}} \\
     & DTI & 0.65\tiny{±0.12} & 0.64\tiny{±0.13} & 0.68\tiny{±0.14} & 0.66\tiny{±0.13} & \textbf{0.69\tiny{±0.06}} & \textbf{0.69\tiny{±0.08}} \\
    \bottomrule
    \end{tabular}
    }
\end{table}%

\subsection{Consideration 2: Adaptive Task Reweighing}

\subsubsection{Challenges.}

Another challenge of cross-dataset brain network analysis is that our previous base model fails to consider relative difficulty of different individual tasks.
From the STT column in Table \ref{tab:fullcomp} we can observe that varying the choice of source task does not lead to uniform improvements on the target performance.
We suspect that this is because some tasks are easier to learn than others, which will converge faster during the meta training phase.
In other words, the base model fails to equally capture the latent knowledge of all source datasets, which potentially hinders the ability of generalization.

\subsubsection{Solutions.}
We first analyze the relationship between the source and target tasks and verify the assumption that source tasks have overlapping knowledge with target tasks.
Then, we propose to improve the meta learning framework so that source tasks could be adaptively optimized during learning.

\begin{figure}
    \centering
    \subfloat[PPMI and BP Task Correlation]{
    \includegraphics[width=0.48\linewidth]{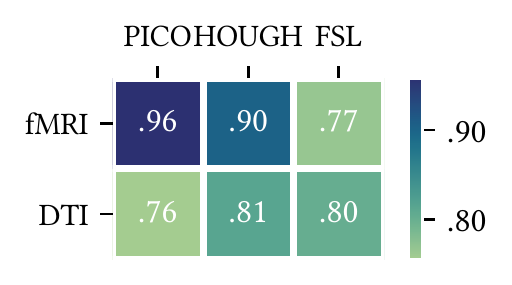}
    \label{fig:bvp}
    }
    \subfloat[PPMI and HIV Task Correlation]{
    \includegraphics[width=0.48\linewidth]{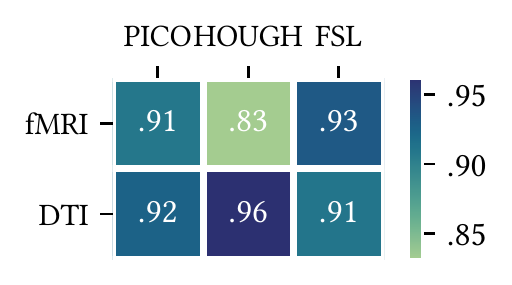}
    \label{fig:hvp}
    }
    \caption{Task correlations with different source and target datasets. We compute the Fisher information estimation derived from the Hessian matrix by training each task using the same architecture as in Section \ref{subsec:trainingeval}. The task embedding is then composed of layer-wise concatenation of the flattened Fisher information matrix.}
\end{figure}

We first investigate the data-level task correlations. In particular, we analyze the task similarity between the HIV and BP modalities (i.e. target) with respect to the PPMI views (i.e. source). Inspired by task2vec \cite{achille2019task2vec}, for each task, we calculate a respective task embedding that stores information regarding its learning difficulty and latent knowledge. In particular, the embedding is derived from the Fisher information estimation of the positive semidefinite upper bound of the Hessian matrix, on which the model is trained on an encoder model using the same objective in Eq.~(\ref{eq:pre-train}).
We visualize the task correlation in cosine similarity among the embeddings on HIV and BP in \cref{fig:bvp,fig:hvp} respectively. It can be seen that there is an inherent correlation among source (i.e. PPMI) and target (i.e. HIV, BP) datasets which indicates that there exists shared properties and latent information among the three categories of brain network data. Our observation can be corroborated with existing clinical research presented in earlier studies \cite{meade2012bipolar, dehner2016parkinsonism, novaretti2016bipolar, pontone2019association, faustino2020risk}, where detailed analyses on the coexistence and co-influence among BP, HIV, and PPMI disease are discussed.
This validates the working effectiveness of our cross-dataset setting since useful and transferable inter-domain knowledge and shared features can be discovered by learning on a source data.
In addition, the visualization also shows a non-uniform task correlation, which suggests that the source tasks are prescribed to varying level of learning and adaptation difficulty relative to the given target task. This demonstrates that the optimizer tends to distribute unequal attention within the source set during meta-training and that the learned initialization will eventually skew towards the optimal of ``easier'' tasks and fails to generalize over ``harder'' tasks. This motivates us to develop dynamic inner-loop optimization rules during meta-training towards an unbiased generalization ability.

Following the mechanism proposed by ALFA \cite{baik2020meta}, during the task-specific inner-loop update, we implement a trainable hyperparameter generator that guides the rate of convergence for the gradient-descent update. The generator processes the learning state as input, which is consisted of a stacked layer-wise value of model parameter and gradient estimate. The generator then outputs a layer-wise learning rate and weight decay coefficient conditioned on the current learning state.
Then, its parameters are updated by the query loss objective as in Eq.~(\ref{eq:pre-train}). Different from the original ALFA, where the encoder parameters are frozen from updating at the outer-loop phase, we allow the encoder to be trainable on the query set for quicker adaptation. We summarize this variant (dubbed MMAR) in Algorithm \ref{alg:alfa}.

\begin{algorithm}[t]
    \centering
    \caption{Multi-task meta-learning with adaptive task reweighing (MMAR)}\label{algorithm2}
    \begin{algorithmic}[1]
        \State \textbf{Input:} meta-train tasks $S_\tau$, meta-test task $T$, encoder $f(\theta)$, hyperparameter generator $g(\phi)$
        \State \textbf{Require:} \text{$\eta$: outer-loop learning rate}
        \State \text{Randomly initialize $\theta$, $\phi$}
        \State $\triangleright$ Meta-training phase
        \While{not done}
        \For{each task $\tau_i$ in $S_\tau$}
        \State Sample $n$ datapoints $\mathcal{D}_i$ from $\tau_i$
        \State \text{Evaluate the gradient} $\nabla_\theta \mathcal{L}_{\mathcal{D}_i}f(\theta)$
        \State Obtain the task-specific learning state $\rho_i = [\nabla_\theta \mathcal{L}_{\mathcal{D}_i}f(\theta), \theta]$
        \State Generate hyperparameters $\alpha$, $\beta$ = $g_\phi(\rho_i)$
        \State Compute the adapted parameters $\theta'_i \gets \beta \odot \theta - \alpha \odot \nabla_\theta \mathcal{L}_{\mathcal{D}_i}f(\theta)$
        \State Sample another set of datapoints $\mathcal{D}'_i$ from $\tau_i$
        \EndFor
        \State Update parameters $\theta \gets \theta - \eta \nabla_\theta \sum_{\mathcal{D}'_i, \theta'_i \sim S_\tau} \mathcal{L}_{\mathcal{D}'_i}f(\theta'_i)$
        \State Update parameters $\phi \gets \phi - \eta \nabla_\phi \sum_{\mathcal{D}'_i, \theta'_i \sim S_\tau} \mathcal{L}_{\mathcal{D}'_i}f(\theta'_i)$
        \EndWhile
        \State Perform $k$-fold evaluation on target tasks
    \end{algorithmic}
    \label{alg:alfa}
\end{algorithm}

\subsubsection{Empirical evaluation.}
In our implementation, the hyperparameter generator network is composed of a two-layer MLP with the input learning state being a concatenation of layer-wise mean of parameter values and gradients. The output of the generator network is a vector containing stacked value of layer-specific learning rate and weight decay coefficients.

Table \ref{tab:alfa} summarizes our experiments on the task adaptive reweighing variant (MMAR). For referential and comparative purpose, the table also includes the performance of the DSL and MML baselines. According to the performance report, MMAR achieves on average 4\% absolute gain in ACC score and 4\% gain in AUC score over the MML implementation. In addition, MMAR outperforms the DSL baseline with a 12\% absolute gain in AUC and 13\% in ACC, which approximates to nearly 21\% relative gain overall. Such improvement indicates that the dynamic inner-loop update adjustment in the meta-training process achieves a more robust approximations to source optimal and a less biased meta initialization. This also leads to faster convergence in target task fine-tuning resulting in an enhanced computational efficiency.

\begin{table}
  \centering
  \caption{Performance with task reweighing techniques.}
  \resizebox{\linewidth}{!}{
    \begin{tabular}{cccccccc}
    \toprule
    \multirow{2}[4]{*}{Dataset} & \multirow{2}[4]{*}{Modality} & \multicolumn{2}{c}{DSL} & \multicolumn{2}{c}{MML} & \multicolumn{2}{c}{MMAR} \\
\cmidrule(lr){3-4} \cmidrule(lr){5-6} \cmidrule(lr){7-8}          &       & AUC & ACC & AUC & ACC & AUC & ACC \\
    \midrule
    \multirow{2}[1]{*}{BP} & fMRI  & 0.55\tiny{±0.11} & 0.54\tiny{±0.14} & 0.62\tiny{±0.08} & 0.62\tiny{±0.10} & \textbf{0.68\tiny{±0.10}} & \textbf{0.66\tiny{±0.08}} \\
          & DTI   & 0.51\tiny{±0.12} & 0.52\tiny{±0.11} & 0.59\tiny{±0.07} & 0.58\tiny{±0.11} & \textbf{0.64\tiny{±0.06}} & \textbf{0.64\tiny{±0.10}} \\
    \midrule
    \multirow{2}[1]{*}{HIV} & fMRI  & 0.63\tiny{±0.18} & 0.64\tiny{±0.12} & 0.69\tiny{±0.10} & 0.70\tiny{±0.09} & \textbf{0.74\tiny{±0.10}} & \textbf{0.76\tiny{±0.08}} \\
          & DTI   & 0.60\tiny{±0.12} & 0.58\tiny{±0.13} & 0.65\tiny{±0.12} & 0.64\tiny{±0.13} & \textbf{0.72\tiny{±0.08}} & \textbf{0.72\tiny{±0.07}}\\
    \bottomrule
    \end{tabular}}
  \label{tab:alfa}
\end{table}

\section{Discussions and Conclusions}

In this work, we first discovered the inherent challenges in learning on small-sized brain network datasets by experimenting under a traditional supervised training baseline. We then formulate this problem into a data-efficient learning objective, where we aim to find a generalizable model initialization that achieves efficient adaption on target tasks. To this end, we leverage transfer learning and meta-learning strategies to serve as backbone frameworks for model pre-training. In addition, we propose an automated atlas transformation design that helps address the incompatibility challenge of cross-dataset brain network ROI template dimensions. We also introduce an adaptive task reweighing algorithm that helps resolve biased learning issues in the conventional meta-training pipeline. Extensive experimentation demonstrated the effectiveness of our proposed methodologies.
It is worth noting that our framework is naturally generic and can be easily scaled to other types of neuroimaging datasets. The training pipeline can also be generalized to any parameterized model that is optimized on different task objectives and data sampling strategies.

Learning on brain network data is still prescribed to various challenges. First, most brain networks are expressed by multiple views and modalities, in which, to achieve a comprehensive feature extraction, would require GNN models to capture complex inter-relations within graph modalities. Simply applying multi-facet meta-learning and separately optimizing on individual views fail to consider the intricacies of some shared and complementary knowledge underneath the multi-view datasets. Second, the target performance on supervised disease classification still suffers from relatively high data variance under the $k$-fold evaluation scheme. This suggests that, assuming given a balanced dataset, the current GNN models are sensitive to batch effects, which would require additional handling of data noise and further development of GNN models that achieve good out-of-distribution performance.

For future investigation, we will primarily focus our work on addressing the aforementioned challenges by performing theoretical and empirical analyses on GNN architectures for brain network learning. We will also undertake to examine the effectiveness of data-efficient learning on neuroimaging data in boarder spectrum of clinical applications. To tackle the data scarcity issue, we will also explore data augmentation and synthetic generation techniques \cite{guo2020systematic, zhu2022survey} to expand available training samples with artificially constructed, domain- and distribution-aware data instances.

\begin{acks}
This research was partly supported by the internal funds and GPU servers provided by the Computer Science Department of Emory University and the University Research Committee of Emory University.
Lifang He was supported by ONR N00014-18-1-2009 and Lehigh's accelerator grant S00010293.
Ying Guo was supproted by National Institute of Health under Award Number R01MH105561.
\end{acks}

\bibliographystyle{ACM-Reference-Format}
\bibliography{kdd22.bib}  

\end{document}